\documentclass{article}
\usepackage{spconf,amsmath,graphicx, bm}

\usepackage{times}
\usepackage{epsfig}
\usepackage{amssymb}
\usepackage{caption}
\usepackage{subcaption}
\usepackage{xcolor}
\usepackage{multirow}
\usepackage{pifont}
\usepackage{adjustbox}
\usepackage{setspace}
\usepackage{mathtools}
\newcommand\tab[1][1cm]{\hspace*{#1}}

\title{SALIENCY DETECTION VIA GLOBAL CONTEXT ENHANCED FEATURE FUSION\\ AND EDGE WEIGHTED LOSS}

\name{Chaewon Park $^{\dagger}$ \thanks{$^{\dagger}$These authors contributed equally} \quad
   Minhyeok Lee $^{\dagger}$ \quad
   MyeongAh Cho \quad
   Sangyoun Lee$^{*}$ \quad}
\address{Yonsei University, Seoul, Republic of Korea}
\begin{document}
%
\maketitle
\begin{abstract}
UNet-based methods have shown outstanding performance in salient object detection (SOD), but are problematic in two aspects. 1) Indiscriminately integrating the encoder feature, which contains spatial information for multiple objects, and the decoder feature, which contains global information of the salient object, is likely to convey unnecessary details of non-salient objects to the decoder, hindering saliency detection. 2) To deal with ambiguous object boundaries and generate accurate saliency maps, the model needs additional branches, such as edge reconstructions, which leads to increasing computational cost. To address the problems, we propose a context fusion decoder network (CFDN) and near edge weighted loss (NEWLoss) function. The CFDN creates an accurate saliency map by integrating global context information and thus suppressing the influence of the unnecessary spatial information. NEWLoss accelerates learning of obscure boundaries without additional modules by generating weight maps on object boundaries. Our method is evaluated on four benchmarks and achieves state-of-the-art performance. We prove the effectiveness of the proposed method through comparative experiments.

\end{abstract}

\begin{keywords}
salient object detection, feature fusion, global context, object boundary, weighted loss
\end{keywords}
\vspace{-0.1cm}
\section{Introduction}
\label{sec:intro}
\vspace{-0.1cm}
Salient object detection (SOD), like human visual attention, aims to detect and segment the most visually prominent objects. SOD is crucial before subsequent vision tasks such as object recognition, tracking, and image parsing.

\begin{figure}[t]
	\setlength{\belowcaptionskip}{-10pt}
	\begin{center}
		\includegraphics[width=1\linewidth]{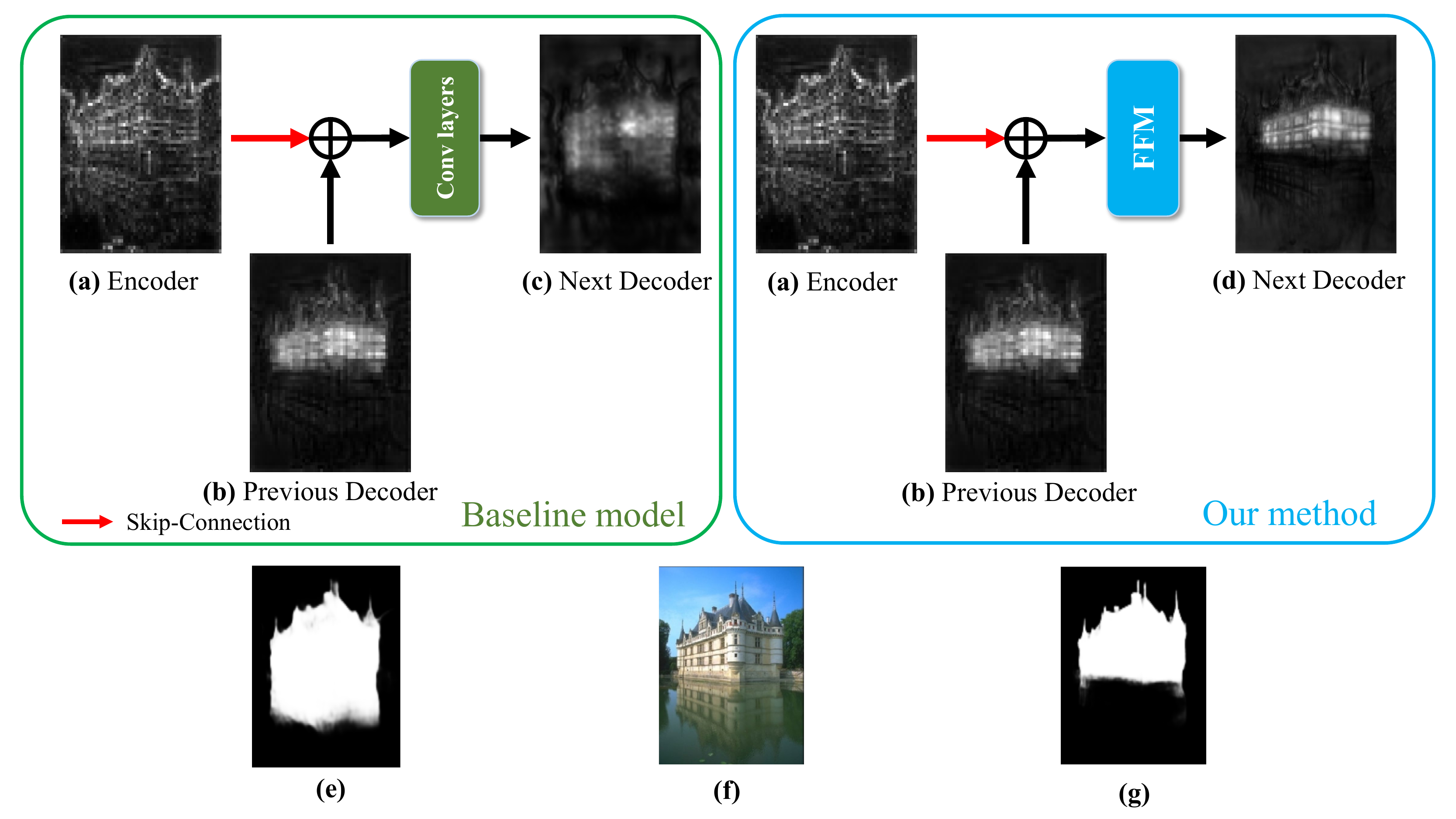}
	\end{center}
	\vspace{-0.5cm}
	\caption{Comparison of baseline model and our method. (f), (a), and (b) are the input, feature of the skip-connected encoder feature, and feature of the previous decoder layer, respectively. (c) and (d) are the decoding results of the skip-connected encoder and previous decoder feature from the baseline model and our method, respectively. (e) and (g) are the segmentation results predicted by the baseline model and our model, respectively.}
	\vspace{-0.1cm}
	\label{fig:1}
\end{figure}

Recently, with the development of deep learning, SOD has achieved rapid progress. In particular, most of the state-of-the-art SOD papers~\cite{hou2017deeply, luo2017non, wang2019salient, liu2019employing} are based on an improved skip-connection-based architecture such as UNet~\cite{ronneberger2015u}. These structures use skip connections to prevent the loss of spatial details and enhance information transfer between the encoder and decoder. However, in the UNet-based model with skip connections, the features between the encoder and the decoder are indiscriminately integrated. As shown in Fig.~\ref{fig:1} (a), the encoder feature has detailed structural information for all objects as well as spatial information for salient objects. However, the decoder feature includes information about salient objects reconstructed from the high-level feature compressed by the encoder, as shown in Fig.~\ref{fig:1} (b). Therefore, there is a large information gap between the encoder and decoder features. Consequently, for an image that contains multiple objects, the encoder's detailed information of non-salient objects is also transmitted to the decoder through a skip connection, acting as serious noise for saliency mask reconstruction. Figs.~\ref{fig:1} (c) and (e) show the result of combining encoder and decoder features of a basic skip-connection-based model (baseline) and the final saliency mask output. Here, it shows that the detailed structural information of the building reflected in the water acts as noise to the decoder, creating an incorrect saliency mask.

\begin{figure*}[t]
	\setlength{\belowcaptionskip}{-10pt}
	\begin{center}
		\includegraphics[width=1\linewidth]{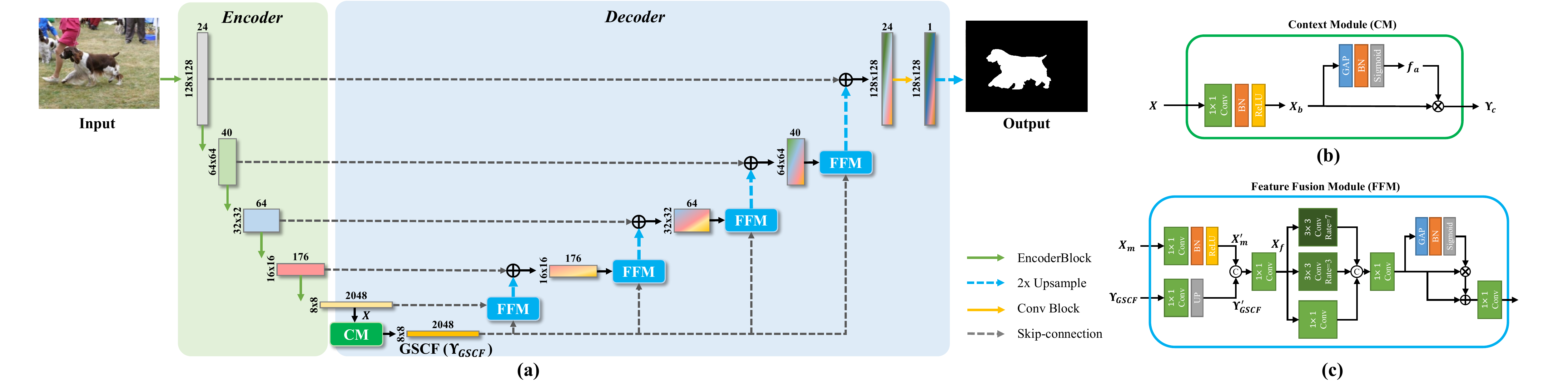}
	\end{center}
	\vspace{-0.5cm}
	\caption{(a) Proposed model architecture. (b) CM structure. (c) FFM structure. (b) is used to augment global context information for the decoder. (c) fuses the encoder, decoder, and global context features.}
	\label{fig:2}
\end{figure*}

Furthermore, many previous works~\cite{wang2019salient, zhao2019egnet, su2019selectivity, wei2020label} have attempted to generate sharp salient object masks by reconstructing edges or boundaries for salient objects. These methods improve performance by providing additional supervision signals to the edges of objects that are difficult to segment. However, they increase the computational cost because additional edge reconstruction branches are added to the model. Moreover, most of these methods~\cite{wang2019salient, zhao2019egnet, su2019selectivity} employ parts of the additional branches—which extract extra information of the edges—during testing, increasing the detection complexity and running time.

In this paper, we propose a novel context fusion decoder network (CFDN) structure to solve these problems. The CFDN consists of the proposed feature fusion module (FFM) and context module (CM). The CM extracts global context information from the encoder and provides important context information to decoder blocks through the FFM. The FFM combines encoder features, decoder features, and global context features generated from the CM. The FFM uses global context information as a guide to reduce the influence of unnecessary spatial information about non-salient objects and helps the decoder to generate an accurate saliency map. As shown in Figs.~\ref{fig:1} (d) and (g), the FFM removes the effect of unwanted structural information of the encoder. Since it is much more difficult to predict ambiguous boundaries, we also propose a novel near edge weighted loss (NEWLoss) function, which provides a weight map to the object boundary by generating distance maps of different features for the background and foreground. Therefore, it efficiently enhances object boundary learning without any additional computational costs or modules, unlike the other methods~\cite{wang2019salient, zhao2019egnet, su2019selectivity, wei2020label} that embed extra edge information and tasks during training and testing.

\vspace{-0.3cm}
\section{PROPOSED METHOD}
\vspace{-0.2cm}
\subsection{Overview}
\vspace{-0.1cm}
The overall model architecture is shown in Fig.~\ref{fig:2} (a). The proposed model consists of an efficientnet-b5~\cite{tan2019efficientnet} encoder and context fusion decoder. The input image passes through the encoder blocks to generate five different resolution feature maps. Among them, the feature of the last encoder block creates a context feature through the CM. In the decoder part, the context feature is upsampled and fused in the FFM with the skip-connected encoder feature and the previous decoder feature. When the final saliency map is generated by passing through the decoding block, the model is learned end-to-end by calculating NEWLoss along with the ground truth.

\vspace{-0.2cm}
\label{sec:method}
\subsection{CFDN}
\noindent
\textbf{CM.} We propose the CM to supplement important context information for each decoder stage. The CM generates the global salient context feature (GSCF) $\Upsilon_{GSCF}$ from the last layer feature $\textbf{X}$ of the encoder. As Fig.~\ref{fig:2} (b) shows, the CM applies global average pooling to $\textbf{X}_b$ to capture global context information and computes an attention vector $\textbf{f}_a$, where $\textbf{X}_b = ConvBNReLU\left(\textbf{X}\right)$ and $ConvBNReLU\left(.\right)$ is a $1 \times 1$ conv-batch normalization-ReLu. Finally, $\textbf{X}_b$ is multiplied by $\textbf{f}_a$ to guide feature learning. As a result, the GSCF contains the core context features for salient objects, which are fed into the context input of the FFM.

\begin{figure}[t]
    \begin{minipage}[b]{1.0\linewidth}
      \centering
      \centerline{\includegraphics[width=1\linewidth]{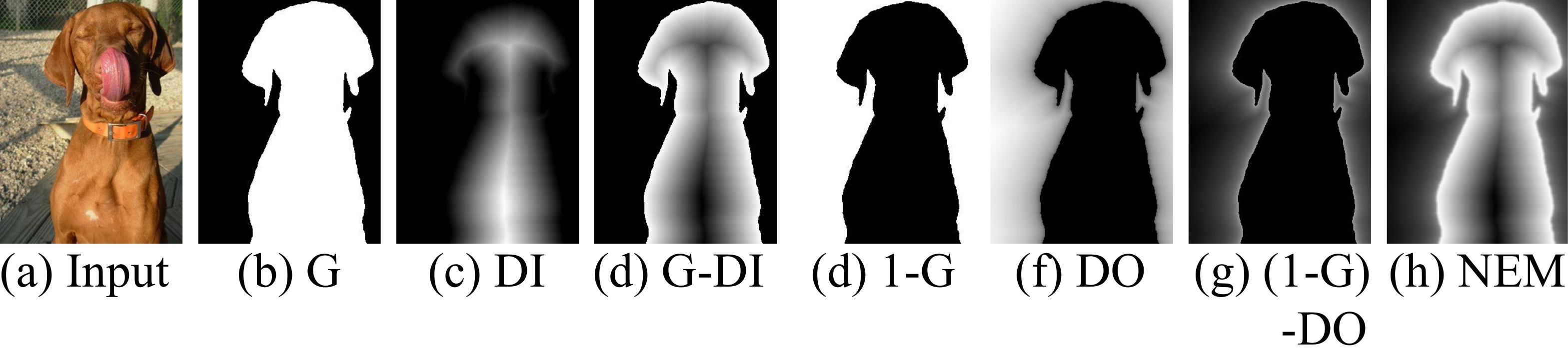}}
    \end{minipage}
    \vspace{-0.7cm}
    \caption{Generation of the $\mathbf{NEM}$. (a), (b), and (e) are the input, ground truth (G), and the reversed ground truth, respectively. (c) and (d) denote the inner distance map (DI) and $G-DI$, respectively. (f) and (g) indicate the outer distance map (DO) and $(1-G)-DO$, respectively. Finally, (h) is the \textbf{NEM}, which is a combination of (d) and (g).}
    \vspace{-0.5cm}
    \label{fig:edgeMask}
\end{figure}

\begin{table*}[t]
  \caption{Performance comparison with state-of-the-art methods on four datasets. The best and the second best are highlighted in \textcolor{red}{red} and \textcolor{blue}{blue} results, respectively.}
  \vspace{-0.2cm}
  \label{table: performance}
  \renewcommand\tabcolsep{2.35pt}
  \renewcommand\arraystretch{1.0}
  \centering
  \footnotesize
  \begin{tabular}{l|cccc|cccc|cccc|cccc}
     \hline
     \hline
     \multirow{2}{*}{\textbf{Algorithm}}  & \multicolumn{4}{c|}{\textbf{ECSSD}~\cite{yan2013hierarchical}} & \multicolumn{4}{c|}{\textbf{PASCAL-S}~\cite{li2014secrets}} & \multicolumn{4}{c|}{\textbf{DUTS-TE}~\cite{wang2017learning}} & \multicolumn{4}{c}{\textbf{DUT-OMRON}~\cite{yang2013saliency}}\\
      
    \cline{2-17}
     & MAE$\downarrow$ & $mF$$\uparrow$ & $\mu F$$\uparrow$ & $S_{\alpha}$$\uparrow$ 
     & MAE$\downarrow$ & $mF$$\uparrow$ & $\mu F$$\uparrow$ & $S_{\alpha}$$\uparrow$ 
     & MAE$\downarrow$ & $mF$$\uparrow$ & $\mu F$$\uparrow$ & $S_{\alpha}$$\uparrow$ 
     & MAE$\downarrow$ & $mF$$\uparrow$ & $\mu F$$\uparrow$ & $S_{\alpha}$$\uparrow$ \\
     \hline
     \hline
     BiDir~\cite{zhang2018bi}         & 0.045 & 0.929 & 0.900 & 0.911 & 0.049 & 0.851 & 0.814 & 0.861 & 0.064 & 0.774 & 0.744 & 0.808 & 0.074 & 0.862 & \textcolor{blue}{0.855} & 0.834 \\
     RAS~\cite{chen2018reverse}           & 0.056 & 0.921 & 0.900 & 0.893 & 0.060 & 0.831 & 0.803 & 0.838 & 0.062 & 0.786 & 0.762 & 0.813 & 0.104 & 0.837 & 0.829 & 0.785 \\
     EGNet~\cite{zhao2019egnet}         & 0.044 & 0.941 & - & 0.913 & 0.043 & 0.880 & - & 0.866 & 0.056 & 0.826 & - & 0.813 & 0.076 & 0.863 & - & 0.848 \\
     AFNet~\cite{feng2019attentive}         & 0.042 & 0.935 & 0.915 & 0.914 & 0.046 & 0.862 & \textcolor{blue}{0.834} & 0.866 & 0.057 & 0.797 & \textcolor{blue}{0.776} & \textcolor{blue}{0.826} & 0.076 & \textcolor{blue}{0.879} & \textcolor{red}{0.866} & 0.841 \\
     BAS~\cite{qin2019basnet}           & 0.037 & 0.942 & - & - & 0.047 & 0.860 & - & - & 0.056 & 0.805 & - & - & 0.076 & 0.854 & - & - \\
     MINet~\cite{pang2020multi}         & \textcolor{blue}{0.036} & 0.943 & \textcolor{red}{0.922} & \textcolor{blue}{0.919} & \textcolor{blue}{0.039} & 0.877 & 0.823 & \textcolor{blue}{0.875} & 0.057 & 0.794 & 0.741 & 0.822 & \textcolor{blue}{0.065} & \textcolor{red}{0.882} & 0.843 & \textcolor{blue}{0.855} \\
     KRN~\cite{xu2021locate}           & \textcolor{blue}{0.036} & \textcolor{blue}{0.946} & - & - & \textcolor{red}{0.034} & \textcolor{red}{0.898} & - & - & \textcolor{blue}{0.049} & \textcolor{blue}{0.827} & - & - & 0.067 & 0.872 & - & - \\
     \hline
     \textbf{Ours} & \textcolor{red}{0.033} & \textcolor{red}{0.951} & \textcolor{blue}{0.916} & \textcolor{red}{0.932} & \textcolor{blue}{0.039} & \textcolor{blue}{0.891} & \textcolor{red}{0.852} & \textcolor{red}{0.894} & \textcolor{red}{0.048} & \textcolor{red}{0.859} & \textcolor{red}{0.821} & \textcolor{red}{0.871} & \textcolor{red}{0.045} & \textcolor{red}{0.882} & 0.845 & \textcolor{red}{0.884} \\
     \hline
     \hline
  \end{tabular}
\end{table*}

\begin{figure*}[t]
	\setlength{\belowcaptionskip}{-10pt}
	\begin{center}
		\includegraphics[width=0.8\linewidth]{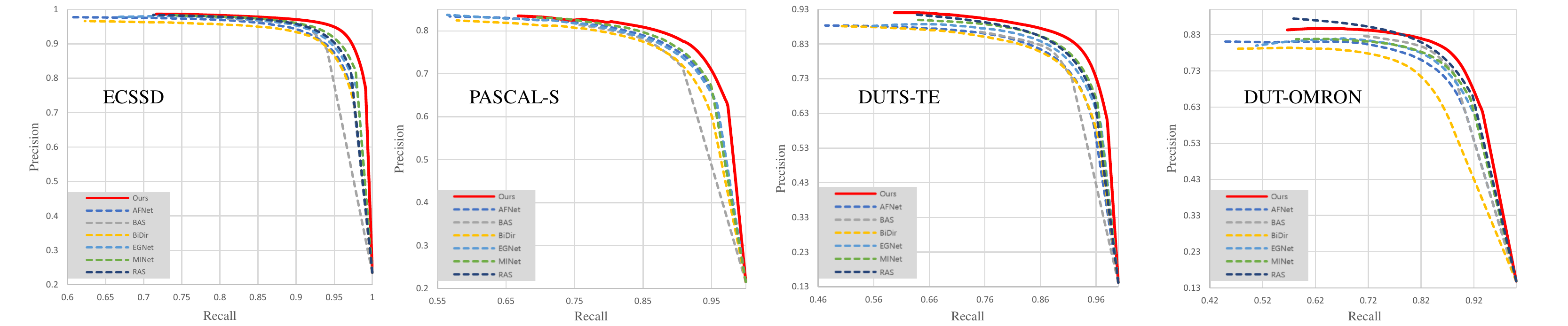}
	\end{center}
	\vspace{-0.5cm}
	\caption{Illustration of PR curves on four datasets.}
	\label{fig:5}
\end{figure*}

\noindent
\textbf{FFM.} We propose the FFM to integrate the encoder and decoder features with different levels of feature representation using global context features as a guide. As shown in Fig.~\ref{fig:2} (a), the first input of the FFM is $\textbf{X}_m = \textbf{X}_e^i + UP\left(\textbf{X}_d^j\right)$, where $\textbf{X}_e^i$ is a skip-connected $i^{th}$ encoder block feature, $\textbf{X}_d^j$ is a $j^{th}$ decoder block feature, and $UP\left(.\right)$ is the upsampling operator. $\textbf{X}_m=X$ only for the first FFM. Since $\textbf{X}_m$ is the sum of two features with different levels of feature representation, we utilize batch normalization to balance the scales of the features, which is expressed as $\textbf{X}_m^{'}=ConvBNReLU\left(\textbf{X}_m\right)$. The second FFM input is the global context feature $\Upsilon_{GSCF}$ generated from the CM. $\Upsilon_{GSCF}$ passes the $1 \times 1$ conv layer and is then upsampled to the same size as $\textbf{X}_m$, which is expressed as $\Upsilon_{GSCF}^{'}=Conv1\left(UP\left(\Upsilon_{GSCF}\right)\right)$, where $Conv1\left(.\right)$ is a $1 \times 1$ conv layer. These two inputs are concatenated to become $\textbf{X}_f$, which is expressed as $\textbf{X}_f = Conv1\left(cat\left(\textbf{X}_m^{'}, \Upsilon_{GSCF}^{'}\right)\right)$. As shown in Fig.~\ref{fig:2} (c), we use parallel convolutional layers with different dilation rates inspired by atrous spatial pyramid pooling (ASPP)~\cite{chen2017rethinking} to capture useful multi-scale spatial information of salient objects and effectively integrate global context information. This layer consists of one $1 \times 1$ conv layer and two $3 \times 3$ conv layers with dilation rates of 3 and 7. This reduces unnecessary spatial information’s influence by using global context information as a guide. As with CM, we use global average pooling in the last FFM step to enhance feature learning.

\vspace{-0.3cm}
\subsection{NEWLoss.}
\label{ssec:loss}
\vspace{-0.2cm}
To generate the predictions with precise boundaries, we propose NEWLoss. First, we make a novel near-edge mask $\mathbf{NEM}\in \mathbb{R}^{1 \times 256 \times 256}$ from the ground truth mask $\mathbf{G} \in \mathbb{R}^{1 \times 256 \times 256}$. Then, we leverage the $\mathbf{NEM}$ as a weight matrix when calculating the pixel-wise loss. Specifically, the $\mathbf{NEM}$ penalizes the large training losses of the pixels near the boundaries of salient objects. \\
\indent The $\mathbf{NEM}$ is formed by merging two distinct masks, where each highlights the inner and outer boundaries of salient objects, respectively. As in~\cite{wei2020label}, we apply the distance transform to the binary $\mathbf{G}$ to get an inner distance map $\mathbf{DI}$ (Fig.~\ref{fig:edgeMask} (c)). By this process, each of the foreground pixel values is converted to the distance between the corresponding pixel and the nearest background pixel. We define the converted pixel values ${DI}_p$ as Eq.~\ref{eq:InD} where $p$ and $q$ represent the pixel of $\mathbf{G}$ and the nearest background pixel, respectively. 
\vspace{-0.1cm}
\begin{equation}
	DI _ { p } = \begin{cases} min \sqrt { (p _ { x }-q _ { x } ) ^ { 2 } +(p _ { y }-q _ { y } ) ^ { 2 } } & , p \in \mathbf{G} _ { fg }  \\ \tab[2.0cm]0 & , p \in \mathbf{G} _ { bg } \end{cases}
\label{eq:InD}
\end{equation}

\noindent Then, we apply the min-max normalization to map the values to $[0, 1]$ and subtract the normalized $\mathbf{DI}$ from $\mathbf{G}$ to make the inner edge map $\mathbf{G-DI}$ (Fig.~\ref{fig:edgeMask} (d)). Therefore, in $\mathbf{G-DI}$, the pixels near the inner boundary of the salient object have large values, while the background and center pixels have the smallest values. This whole transformation from $\mathbf{G}$ to $\mathbf{G-DI}$ is expressed as the following function $EdgeTF(\mathbf{DI}, p)$:
    \vspace{-0.15cm}
	\begin{equation}
		EdgeTF(\mathbf{DI}, p) = p- \frac{ DI _ { p } -min(\mathbf{DI}) } { max(\mathbf{DI})-min(\mathbf{DI}) }
	\end{equation}

Subsequently, we make an inverse ground truth $\mathbf{1-G}$ (Fig.~\ref{fig:edgeMask} (e)) and then, repeat the same process as Eq.~\ref{eq:InD} to make an outer distance map $\mathbf{DO}$ (Fig.~\ref{fig:edgeMask} (f), Eq.~\ref{eq:OD}). 
    \begin{equation}
    \footnotesize
		{DO _ { p } = \begin{cases} min \sqrt { (p _ { x }-q _ { x } ) ^ { 2 } +(p _ { y }-q _ { y } ) ^ { 2 } } & , p \in \mathbf{1-G} _ { fg }  \\ \tab[2.0cm]0 & , p \in \mathbf{1-G} _ { bg } \end{cases}}
	\label{eq:OD}
	\end{equation}
	
\noindent Next, we operate $EdgeTF(\mathbf{DO}, p)$ to acquire the outer edge map $\mathbf{(1-G)-DO}$ (Fig.~\ref{fig:edgeMask} (g)) where the background pixel values are large near the outer boundary of the salient object, while the values in the the inner part of the object are the smallest. Finally, we obtain the $\mathbf{NEM}$ (Fig.~\ref{fig:edgeMask} (h)) by adding $\mathbf{G-DI}$ and $\mathbf{(1-G)-DO}$. \\
\indent The $\mathbf{NEM}$ is then combined with the binary cross entropy loss~\cite{de2005tutorial}. It is used to amplify the loss values near the boundary pixels by multiplying the edge-emphasized weights. Our NEWLoss function is defined as:
    \vspace{-0.15cm}
    \begin{equation}
         \mathbf{BCE}(p, \widehat{ p } ) = p \times \log \widehat{ p } +(1-p ) \times \log(1- \widehat{ p } )
    \end{equation}
    \vspace{-0.3cm}
    \begin{equation}
        \begin{split}
        \footnotesize
		 NEWLoss(\mathbf{G}, \mathbf{\widehat{ G }} ) &= - \frac{ 1 } { N \times M } \sum _ { x=1 } ^ { N } \sum _ { y=1 } ^ { M } [(\mathbf{NEM} _ { (x,y) }+\eta) \\ &\quad\quad \times \mathbf{BCE}(p _ { (x,y) }, \widehat{ p } _ { (x,y) } )]
		\end{split}
		\label{eq:nembce}
	\end{equation}
    
    
    
    
\vspace{-0.1cm}
\noindent where $\mathbf{\widehat{ G }}$ is the predicted output of our network. Furthermore, $p _ { (x,y) }$ and $\widehat{ p } _ { (x,y) }$ indicate each pixel of $\mathbf{G}$ and $\mathbf{\widehat{ G }}$ at location $(x,y)$, respectively. Additionally, $N$ and $M$ are the number of pixels in the width and height axis, respectively. In Eq.~\ref{eq:nembce}, we add a value $\eta$ to the $\mathbf{NEM}$ to avoid zero values.

\begin{figure}[t]
      \centering
      \centerline{\includegraphics[width=1.0\linewidth]{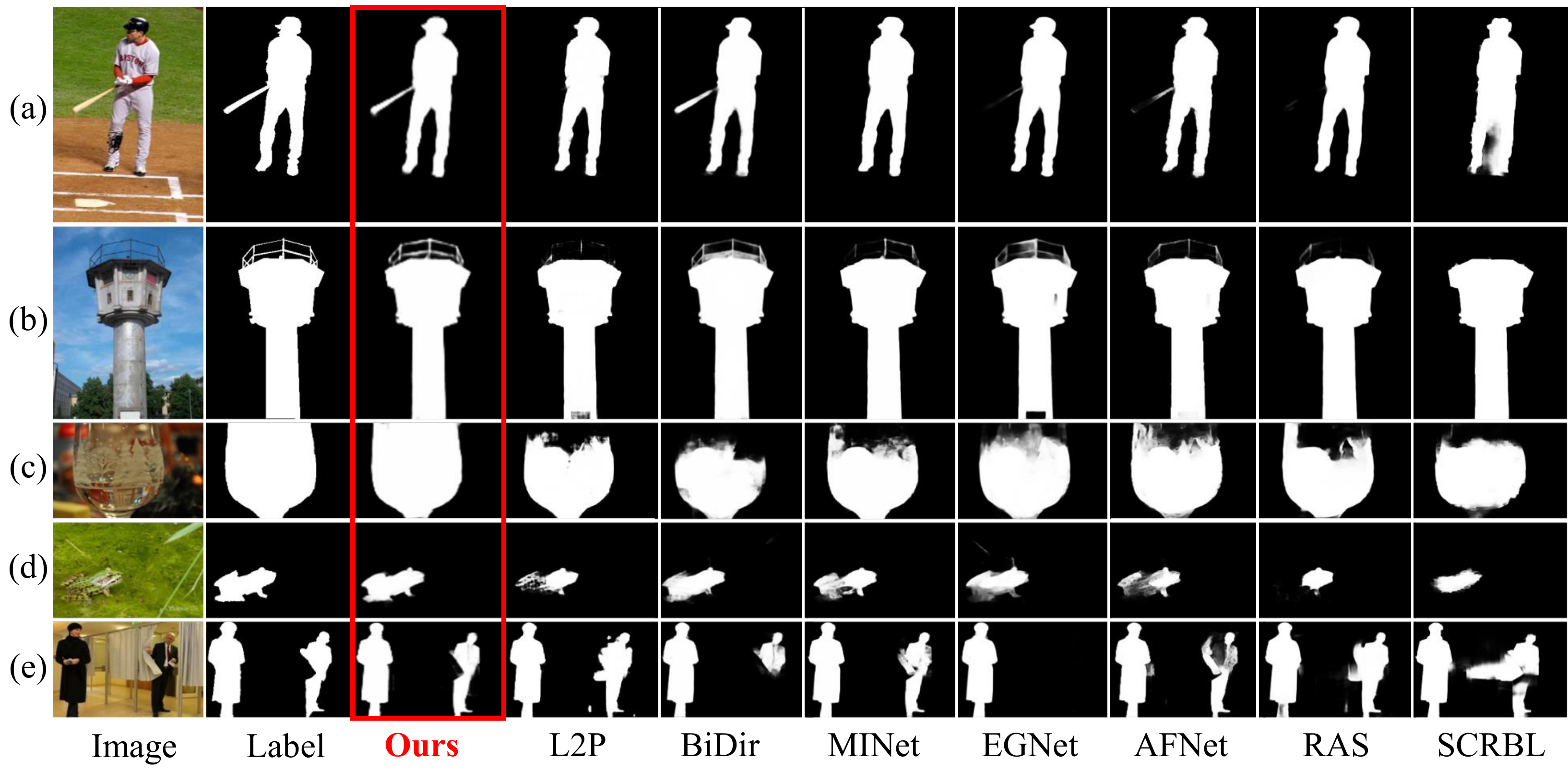}}
    \vspace{-0.3cm}
    \caption{Qualitative comparison on various challenging samples.}
    \vspace{-0.3cm}
    \label{fig:output_compar}
\end{figure}

\vspace{-0.2cm}
\section{Experiment}
\label{sec:experiment}
\vspace{-0.1cm}
\subsection{Experimental Setup}
\vspace{-0.1cm}
{\noindent \bf SOD datasets.}
We evaluate our model on four popular benchmarks: ECSSD~\cite{yan2013hierarchical}, PASCAL-S~\cite{li2014secrets}, DUT-OMRON~\cite{yang2013saliency}, and DUTS~\cite{wang2017learning}. Among these, DUTS~\cite{wang2017learning} is the largest dataset and contains two subsets: DUTS-TR with $10553$ training images and DUTS-TE with $5019$ test images. Following the protocol of most previous studies~\cite{liu2019employing, zhao2019egnet, wei2020label, pang2020multi, zhou2020interactive}, we train our network on DUTS-TR and test on the other datasets for fair comparison.

{\noindent \bf Evaluation metrics.}
As in previous works~\cite{hou2017deeply, wang2019salient, liu2019employing, zhao2019egnet, wei2020label, zhou2020interactive, pang2020multi}, we adopt the four widely used measures for our evaluation: mean absolute error ($MAE$)~\cite{borji2015salient}, maximum F-measure ($mF$), mean F-measure ($\mu F$), and structure measure ($S_{\alpha}$)~\cite{fan2017structure}. A smaller $MAE$ and larger $mF$, $\mu F$, and $S_{\alpha}$ indicate better detection performance. We also demonstrate the precision-recall (PR) curve to show the overall performance.

{\noindent \bf Implementation details.}
Our experiments are all implemented in PyTorch, with a single Nvidia GeForce RTX 3090. We utilize the Adam optimizer ($weight\:decay=10^{-2}$ and $eps=10^{-3}$) to train our network. Furthermore, the initial learning rate is set to $10^{-4}$, which is gradually decreased to $10^{-5}$ by using a cosine annealing scheduler. The parameter $\eta$ is set to 1. We train our model on DUTS-TR~\cite{wang2017learning} for $100$ epochs with a batch size of $28$. All the training images are loaded in RGB scale, and the intensity of the pixels is normalized to $[-1,1]$. Then, they are resized to $300\times300$ and randomly cropped to $256\times256$. The horizontal flip and the vertical flip are applied for data augmentation. Additionally, we adopt EfficientNet-b5~\cite{tan2019efficientnet}, pre-trained on ImageNet, as our network's backbone. 

\vspace{-0.2cm}
\subsection{Comparison with State-of-the-art Methods}
\label{sec:comparison}
\vspace{-0.1cm}
{\noindent \bf Quantitative comparison.}
As Table~\ref{table: performance} shows, our model outperforms or reaches various state-of-the-art methods without any extra supervisions. In particular, our work achieves a big performance improvement in DUTS-TE~\cite{wang2017learning}, outperforming all the counterpart models in all four measures. It is also worth noting that our $S_{\alpha}$ score is the largest in all datasets. The large $S_{\alpha}$ corresponds to the high similarity in the global structure of the segmentation map, mirroring the effect of our CM and FFM. Additionally, we present the PR curves in Fig.~\ref{fig:5}, in which ours lie above all the others.

{\noindent \bf Qualitative comparison.}
In Fig.~\ref{fig:output_compar}, we compare our predicted saliency maps with those of the seven recently proposed networks: L2P~\cite{zeng2018learning}, BiDir~\cite{zhang2018bi}, MINet~\cite{pang2020multi}, EGNet~\cite{zhao2019egnet}, AFNet~\cite{feng2019attentive}, RAS~\cite{chen2018reverse}, and SCRBL~\cite{zhang2020weakly}. Our network is capable of segmenting thin edges (Fig.~\ref{fig:output_compar} (b)) and effective for objects whose boundaries are difficult to distinguish from the backgrounds due to similarities in colors and patterns, and low frequency (Figs.~\ref{fig:output_compar} (c), (d)). Moreover, it is robust to complex scenes where multiple objects appear and their correlations are important (Figs.~\ref{fig:output_compar} (a), (e)). 

{\noindent \bf Ablation study.} 
We conduct ablation studies to validate the impact of our CFDN and NEWLoss. Table~\ref{table:ablation} illustrates the experimental results on different combinations of the two components. Clearly, both the CFDN and NEWLoss function contribute considerably to the performance enhancement. 

\begin{table}
    \centering
    \caption{Performance with different combinations of our contributions. R50 and R101 are ResNet50 and ResNet101~\cite{he2016deep}, respectively. A and B denote NEWLoss and the CFDN, respectively.}
    \label{table:ablation}
    \renewcommand\tabcolsep{1.0pt}
    \renewcommand\arraystretch{1}
    \footnotesize
    \begin{tabular} {c|c|c|cccc|cccc}
    \hline
    \hline
    {\textbf{Back-}}  & \multirow{2}{*}{\textbf{A}} & \multirow{2}{*}{\textbf{B}} &\multicolumn{4}{c|}{\textbf{ECSSD}~\cite{yan2013hierarchical}} & \multicolumn{4}{c}{\textbf{DUTS-TE}~\cite{wang2017learning}}\\
         \cline{4-11}
         {\textbf{bone}}&  &  & MAE$\downarrow$ & $mF$$\uparrow$ & $\mu F$$\uparrow$ & $S_{\alpha}$$\uparrow$ 
         & MAE$\downarrow$ & $mF$$\uparrow$ & $\mu F$$\uparrow$ & $S_{\alpha}$$\uparrow$ \\\hline\hline
        \multirow{4}{*}{\textbf{R50}} &  &  & 0.042 & 0.938 & 0.910 & 0.918 & 0.042 & 0.882 & 0.852 & 0.888 \\
         & {\ding{51}} &  & 0.040 & 0.940 & 0.912 & 0.920 & 0.041 & 0.883 & 0.854 & 0.890 \\ 
         &  & {\ding{51}} & 0.039 & 0.940 & 0.914 & 0.921 & 0.040 & 0.884 & 0.855 & 0.889 \\
         & {\ding{51}} & {\ding{51}} & \textbf{0.037} & \textbf{0.942} & \textbf{0.918} & \textbf{0.922} & \textbf{0.039} & \textbf{0.885} & \textbf{0.859} & \textbf{0.891} \\ \hline
        \multirow{4}{*}{\textbf{R101}} &  &  & 0.039 & 0.942 & 0.915 & 0.925 & 0.045 & 0.871 & 0.841 & 0.877 \\
         & {\ding{51}} &  & 0.038 & 0.943 & 0.917 & 0.926 & 0.043 & 0.876 & 0.846 & 0.879 \\ 
         &  & {\ding{51}} & 0.037 & 0.944 & 0.917 & 0.926 & 0.042 & 0.878 & 0.849 & 0.882 \\
         & {\ding{51}} & {\ding{51}} & \textbf{0.035} & \textbf{0.945} & \textbf{0.920} & \textbf{0.928} & \textbf{0.040} & \textbf{0.881} & \textbf{0.856} & \textbf{0.885} \\
    \hline
    \hline
    \end{tabular}
    \vspace{-0.3cm} 
\end{table}

\vspace{-0.2cm}
\section{CONCLUSION}
\label{sec:conclustion}
\vspace{-0.2cm}
In this paper, we aimed to detect salient objects by alleviating the indiscriminate feature integration problem of the FCN-based methods. Specifically, we proposed a novel CFDN that combines encoder and decoder features of large feature representation gaps, in a way that makes up for their different distributions. Therefore, the CFDN supplements the insufficient contextual information while producing an output saliency map. Furthermore, we proposed the NEWLoss function to enhance the learning of edge pixels without any additional modules. Thus, the CFDN trained with NEWLoss achieved state-of-the-art performance in four benchmarks. Extensive experiments clearly demonstrated the effectiveness of our method.

\newpage
\begingroup
\setstretch{0.95}
\bibliographystyle{IEEEbib}
\bibliography{ref}
\endgroup

\end{document}